\documentclass{article}

\PassOptionsToPackage{numbers, compress}{natbib}
 \usepackage[preprint]{neurips_2026}
\usepackage[utf8]{inputenc} 
\usepackage[T1]{fontenc}    
\usepackage{hyperref}       
\usepackage{url}            
\usepackage{booktabs}       
\usepackage{amsfonts}       
\usepackage{nicefrac}       
\usepackage{microtype}      
\usepackage{xcolor}         

\usepackage{amsmath,amsfonts,bm}

\def\eqref#1{equation~\ref{#1}}

\def\1{\bm{1}}

\DeclareMathAlphabet{\mathsfit}{\encodingdefault}{\sfdefault}{m}{sl}
\SetMathAlphabet{\mathsfit}{bold}{\encodingdefault}{\sfdefault}{bx}{n}

\usepackage{graphicx}
\usepackage{subcaption}
\usepackage{caption}
\usepackage[capitalize,noabbrev]{cleveref}
\usepackage{duckuments}
\usepackage{multirow}
\usepackage{makecell}
\usepackage{enumitem}
\usepackage{colortbl}
\usepackage{xspace}
\usepackage{lipsum}
\usepackage{amsmath}
\usepackage{amssymb}
\usepackage{mathtools}
\usepackage{amsthm}
\usepackage{bbm}
\usepackage{listings}
\usepackage{wrapfig}
\usepackage{float}
\usepackage{minted}

\usepackage{algorithm}
\usepackage{algpseudocode}
\usepackage{bm}
\usepackage{blindtext}
\usepackage{adjustbox}
\usepackage{soul}
\usepackage{tcolorbox}
\usepackage[normalem]{ulem}
\useunder{\uline}{\ul}{}
\usepackage{pifont}
\newcommand{\cmark}{{\ding{51}}}
\newcommand{\xmark}{{\ding{55}}}

\lstdefinelanguage{yaml}{
  keywords={true,false,null},
  basicstyle=\ttfamily\small,
  keywordstyle=\color{blue},
  commentstyle=\color{gray},
  stringstyle=\color{black},
  breaklines=true,
  breakatwhitespace=true,
  showstringspaces=false
}

\ifcsname DeclareUnicodeCharacter\endcsname
  \DeclareUnicodeCharacter{221E}{\ensuremath{\infty}}
\fi

\crefformat{table}{Table~#2#1#3}
\crefformat{figure}{Figure~#2#1#3}
\crefformat{section}{Section~#2#1#3}
\crefformat{appendix}{Appendix~#2#1#3}
\crefformat{equation}{Eq.~#2(#1)#3}

\definecolor{RowHighlight}{gray}{0.9}

\theoremstyle{plain}

\theoremstyle{definition}

\theoremstyle{remark}

\usepackage[textsize=tiny]{todonotes}

\newcommand{\mname}{ExComm\xspace}

\title{ExComm: Exploration-Stage Communication for Error-Resilient Agentic Test-Time Scaling}

\makeatletter
\def\blfootnote{\xdef\@thefnmark{}\@footnotetext}
\makeatother

\author{%
  Woomin Song$^{1}$ 
  \quad Beomjun Kim$^{1}$
  \quad Daewon Choi$^{1}$
  \quad Sai Muralidhar Jayanthi$^{2}$ \\
  \bfseries \quad Saket Dingliwal$^{3\dagger}$ 
  \quad Jinwoo Shin$^{1}$ \quad Aram Galstyan$^{2}$ \\[4pt]
  \mdseries $^{1}$KAIST \qquad $^{2}$Amazon AGI \qquad $^{3}$Together AI
}

\begin{document}

\maketitle
\blfootnote{\ignorespaces$^{\dagger}$Work done at Amazon.}

\begin{abstract}
A common failure mode in long-horizon agentic test-time scaling is error propagation, where factual errors or invalid deductions introduced at intermediate steps persist in the agent's belief state and contaminate later reasoning. Existing test-time scaling methods provide limited control over this process, as they often rely on agents to detect their own mistakes, select among flawed trajectories, or refine solutions only after errors have already shaped the reasoning path. 
We propose \mname, a communication protocol for exploration-stage agentic test-time scaling. \mname is motivated by the empirical observation that the majority of intermediate errors in parallel agentic reasoning produce detectable cross-agent factual conflicts. Leveraging the iterative structure of agentic workflows, \mname periodically audits agent belief states to detect such conflicts, resolves them through a dedicated tool-based verification loop, and returns concise, targeted feedback to the involved agents.
Corrections are incorporated through soft belief updates, which append verified feedback rather than overwriting existing beliefs. Furthermore, to prevent collapsing trajectory diversity due to communication, \mname further introduces a trajectory diversification module that redirects redundant trajectories toward orthogonal strategies. Experiments on AIME 2024, AIME 2025, and GAIA with Gemini-2.5-Flash-Lite and Qwen3.5-4B show that \mname consistently outperforms strong test-time scaling baselines, achieving average performance gains of 5.7\% and 5.0\% over the best-performing baselines, respectively. Further analyses demonstrate improved error recovery, favorable scaling behavior, stronger diversity than adapted communication baselines, and the best performance-cost trade-off among the evaluated methods.
\end{abstract}

\section{Introduction}
\label{sec:intro}
Test-time scaling has emerged as a powerful approach for improving LLM performance without additional training \citep{snell2024scaling}. By allocating additional computation at inference time, methods such as sequential revision \citep{madaan2023self, lee2025evolving}, parallel sampling strategies including Best-of-$N$ and Self-Consistency \citep{chen2023universal,brown2024large}, and tree search techniques \citep{yao2023tree,dvts} have achieved substantial gains across reasoning benchmarks. Recently, this paradigm has been extended to agentic workflows \citep{zhu2025scaling, zhang2025co}, where models solve problems through multiple rounds of generation, tool use, and interaction with an environment.

\begin{figure*}[th]
\centering\small
\centering
\includegraphics[width=\linewidth]{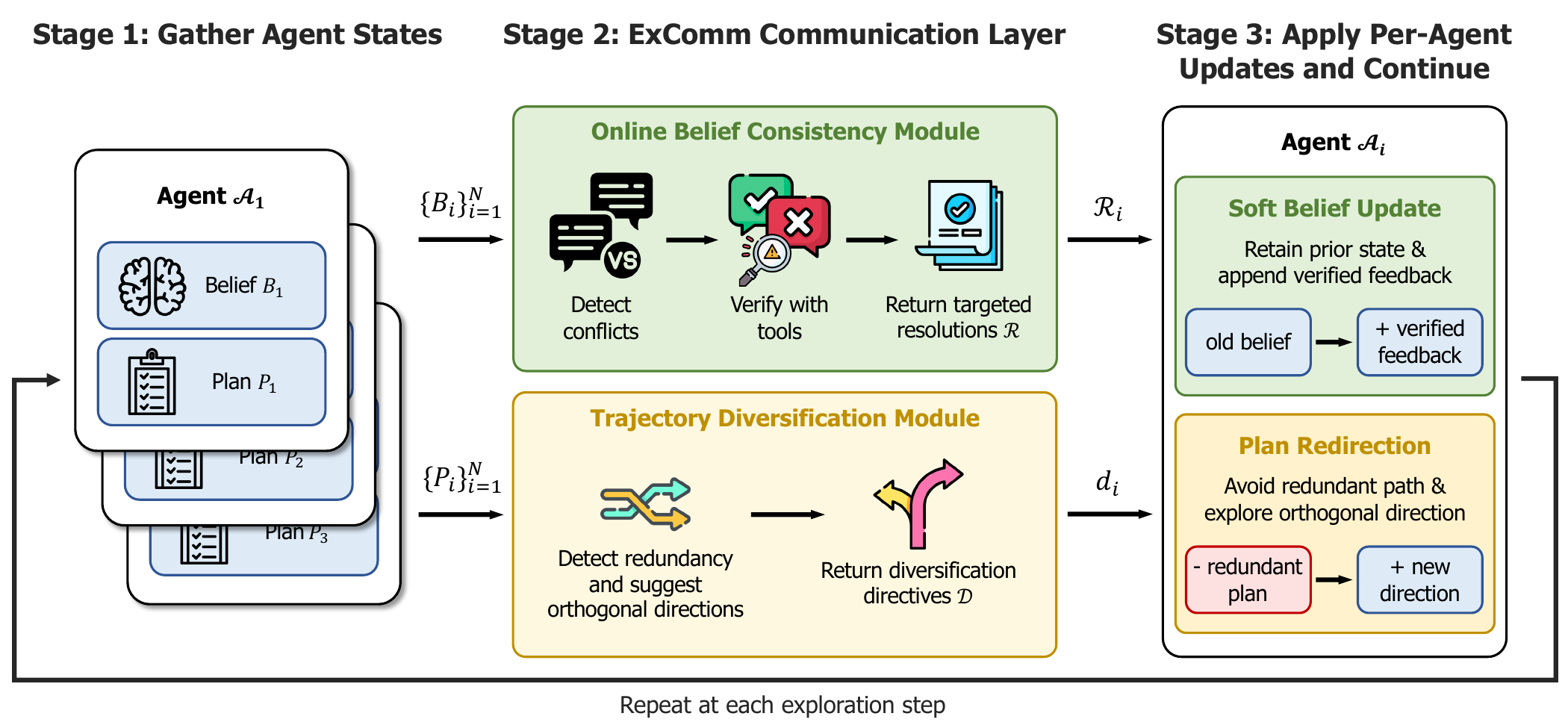} 
\vspace{-2mm}
\caption{
\textbf{Overview of \mname.}
\mname augments a standard agentic test-time scaling loop with an exploration-stage communication step. After each execution step, the Online Belief Consistency Module gathers agent belief states $\{B_i\}_{i=1}^{N}$, detects factual conflicts, resolves them through tool-augmented verification, and produces a set of targeted resolutions $\mathcal{R}$. Each agent $\mathcal{A}_i$ receives only the relevant subset $\mathcal{R}_i \subseteq \mathcal{R}$, which is applied as a soft belief update by appending verified feedback without overwriting the prior belief state. In parallel, the Trajectory Diversification Module analyzes agent plans $\{P_i\}_{i=1}^{N}$, identifies redundant strategies, and produces a set of diversification directives $\mathcal{D}$. When a directive is assigned to agent $\mathcal{A}_i$, it receives $d_i \in \mathcal{D}$ and updates its plan to encourage orthogonal exploration. The updated agents then continue the base execution loop at the next exploration step.
}
\label{fig:concept}
\vspace{-2mm}
\end{figure*}

A common failure mode in long-horizon problem solving is error propagation, where factual errors or invalid deductions introduced at intermediate steps persist in the agent's belief and contaminate later decisions. Existing test-time scaling methods provide limited control over this process. Parallel sampling relies heavily on agents detecting their own mistakes, which is often unreliable \citep{huang2023large,kamoi2024can}. Tree search mitigates errors through branch selection rather than explicit correction. Therefore, errors that are not pruned can continue to shape downstream reasoning.
Post-hoc aggregation or refinement methods intervene only after errors have already shaped the 
reasoning trajectory.

At the same time, agentic workflows provide a natural point of intervention for error control. Unlike static reasoning, they involve dynamic interaction with an environment, where agents alternate between reasoning and actions such as tool calls while updating intermediate beliefs over multiple rounds. This iterative structure creates a natural opportunity for early error detection and correction during the exploration stage of agentic test-time scaling, where agents are still gathering evidence and forming their reasoning trajectories, before candidate solutions are created. We hypothesize that structured communication among parallel reasoning trajectories at this stage can help detect and resolve belief-level errors before they propagate, ultimately improving final answer quality.

To this end, we propose \mname, an exploration-stage communication protocol for agentic test-time scaling. 
\mname is motivated by the empirical observation that the majority of intermediate errors in parallel agentic reasoning introduce cross-agent factual conflicts (67–71\% across benchmarks, see \Cref{sec:method-factcheck}), making them detectable without relying on self-correction.
\Cref{fig:concept} illustrates the core components of \mname. When a conflict is detected, the Online Belief Consistency Module resolves the discrepancy using its own tool-based verification loop. The resulting correction is then returned to the involved agents as concise, targeted feedback. Each agent applies a soft belief update, appending the correction to its belief state rather than overwriting the original belief. This allows agents to benefit from verified information while retaining the ability to recover from potential verifier errors.

A key challenge in exploration-stage communication is the risk of diversity collapse. Excessive information sharing can prematurely synchronize reasoning trajectories, undermining the benefits of parallel scaling. To avoid this, the consistency module provides targeted feedback that addresses only the specific inconsistency, rather than a broad summary of other agents' beliefs. \mname further incorporates a trajectory diversification module, which monitors high-level strategies across agents and encourages orthogonal directions when trajectories become overly similar. By combining targeted belief correction with explicit diversity preservation, \mname mitigates error propagation while maintaining the benefits of parallel exploration.

We evaluate \mname on challenging tool-based reasoning setups, including AIME 2024 and AIME 2025 \citep{balunovic2025aime} with code interpreter access, and GAIA \citep{mialon2023gaia} with code interpreter, file management tools, and web search. Experiments on Gemini-2.5-Flash-Lite \citep{comanici2025gemini25} and Qwen3.5-4B \citep{qwen35} show that \mname consistently outperforms strong test-time scaling baselines, including sequential revision, parallel scaling, and test-time tree search. \mname achieves average performance gains of 5.7\% and 5.0\% over the best-performing baselines with the two models, respectively. Further analysis shows that \mname substantially improves error recovery during reasoning, supporting our claim that exploration-stage communication helps agents correct erroneous beliefs before they propagate.

Additional analyses show that \mname scales with both model size and the number of agents, consistently outperforming baselines across settings. 
We also compare against adapted communication protocols, including Multi-Agent Debate \citep{du2024mad} and Mixture-of-Agents \citep{wang2024moa}, and show that \mname achieves stronger accuracy while preserving higher trajectory diversity.
Ablation studies validate the contributions of both the consistency module and the trajectory diversification module to accuracy and error recovery. Finally, efficiency analysis shows that \mname achieves the best performance-cost trade-off among the evaluated methods, making it a practical approach for agentic test-time scaling.

\section{Related Works}
\label{sec:related-works}

\textbf{LLM Agents.}
Large Language Models (LLMs) have evolved from static text generators into agents capable of solving complex, real-world tasks \citep{react, shen2023hugginggpt, wang2024openhands}. By integrating tools such as calculators, search engines, and APIs, agentic systems can extend standalone models with capabilities such as precise computation and real-time information retrieval \citep{schick2023toolformer, yuan2025easytool}. Unlike static reasoning, agentic workflows are often implemented through a ReAct loop \citep{react}, alternating between reasoning and tool use. This iterative structure provides a natural intervention point during exploration, before a final answer is produced, which \mname uses for error recovery.

\textbf{Test-Time Scaling.}
Test-time scaling (TTS) improves model performance by allocating additional computation during inference without modifying model parameters \citep{snell2024scaling}. Common approaches include parallel sampling \citep{chen2023universal, brown2024large}, sequential revision \citep{madaan2023self, lee2025evolving}, and search-based methods \citep{yao2023tree, tian2024toward, dvts}. While most work studies TTS in static, text-based settings, its application to agentic workflows remains relatively underexplored \citep{zhu2025scaling}. We study an agentic TTS framework focused on preventing error propagation.

One line of TTS research improves reasoning through communication among multiple LLM agents \citep{du2024mad, wang2024moa}. Representative methods such as Multi-Agent Debate and Mixture-of-Agents iteratively exchange, aggregate, or refine intermediate outputs to improve candidate solutions. 
\mname is designed for a different problem: exploration-stage error control. Rather than refining candidate solutions, \mname detects factual conflicts during execution, resolves them through tool-based verification, and returns concise, targeted feedback before errors propagate.

\section{\mname}
\label{sec:method}

In this section, we present the architecture of \mname. We begin in \Cref{sec:method-base} by formalizing a base agent loop that serves as the minimal substrate on top of which our framework is built.
We then introduce the two core modules of \mname. 
In \Cref{sec:method-factcheck}, we describe an online belief consistency module that detects and resolves factual inconsistencies across parallel agents, preventing erroneous beliefs from propagating through subsequent reasoning steps. In \Cref{sec:method-diversifier}, we present a trajectory diversification module that jointly analyzes agents’ plans and selectively redirects redundant trajectories toward orthogonal reasoning strategies. 
Together, these modules allow the agents to detect and recover from errors early-on, preventing the error from propagating and contaminating the later reasoning stages. An overview of the resulting control flow is provided in \Cref{fig:concept}.

\subsection{Base Agent Loop}
\label{sec:method-base}

Before introducing our approach, we first formalize the problem setup. Motivated by recent progress in plan-based execution and memory-based agents, we consider an agent that maintains explicit plans and a structured belief. Formally, a stateful problem-solving agent is defined as
\[
\mathcal{A} := (B, P, \mathcal{T}),
\]
where $B$, $P$, and $\mathcal{T}$ denote the agent’s belief, plan, and available toolkit, respectively.

\begin{wrapfigure}{r}{0.45\textwidth}
\centering
\begin{minipage}{0.45\textwidth}
\vspace{-1em}
\begin{algorithm}[H]
\caption{Base Agent Loop}
\label{alg:base_agent}
\small
\begin{algorithmic}

\Procedure{InitializeAgent}{$p, \mathcal{T}$}
    \State $B \gets \textsc{InitBeliefState}(p)$
    \State $P \gets \textsc{PlanInit}(p, B)$
    \State \Return{$B, P, \mathcal{T}$}
\EndProcedure

\Procedure{AgentStep}{$B, P, \mathcal{T}$}
    \State \texttt{\color{gray} /* Deliberation */}
    \State $t \gets \textsc{SelectTask}(P, B)$
    \State \texttt{\color{gray} /* Execution Loop (ReAct) */}
    \State $\xi_t \gets \textsc{ExecuteTask}(t, B, \mathcal{T})$
    \State \texttt{\color{gray} /* Belief update */}
    \State $B' \gets \textsc{UpdateBelief}(B, \xi_t)$
    \State \texttt{\color{gray} /* Optional replanning */}
    \State $P' \gets \textsc{Replan}(P, B', \xi_t)$
    \State \Return{$B', P'$}
\EndProcedure

\Procedure{AgentLoop}{$p, \mathcal{T}$}
    \State $B, P, \mathcal{T} \gets \textsc{InitializeAgent}(p, \mathcal{T})$
    \While{\textbf{not} \textsc{Terminate}($P, B$)}
        \State $B, P \gets \textsc{AgentStep}(B, P, \mathcal{T})$
    \EndWhile
    \State \Return{$B$}
\EndProcedure

\end{algorithmic}
\end{algorithm}
\end{minipage}
\vspace{-2em}
\end{wrapfigure}

The belief $B$ is a compact document containing factual information available to the agent about the problem, including the problem specification itself as well as information acquired during the problem-solving process. The plan $P$ is a to-do list that records both executed steps and remaining planned actions. The toolkit $\mathcal{T}$ represents the set of tools available to the agent during execution.

\Cref{alg:base_agent} illustrates the agent loop used throughout this work. Given a problem instance $p$, the agent first initializes its belief state $B$ using the information provided in the problem, and then constructs an initial execution plan based on $p$ and $B$. The agent subsequently proceeds through a sequence of execution steps, iteratively processing tasks corresponding to individual items in the plan.

At each execution step, the agent selects a task $t$ from the current plan and executes it using a ReAct-style loop with access to the toolkit $\mathcal{T}$. Upon completion of the task, the agent updates its belief $B$ using the resulting execution log $\xi_t$. The plan $P$ is then optionally revised based on the updated belief and the execution log. We refer to the resulting sequence of execution steps as a \emph{reasoning trajectory}.

In the following sections, we describe the core components of \mname in detail. The proposed modules are applied on the parallel problem-solving agents after they finish each execution step.

\subsection{Online Belief Consistency Module}
\label{sec:method-factcheck}

Long-horizon agentic reasoning is vulnerable to error propagation, where an incorrect intermediate claim, once internalized, impacts all subsequent reasoning steps. Motivated by the intuition that detecting conflicts across multiple reasoning trajectories is substantially easier than identifying errors within a single trajectory, we introduce the Online Belief Consistency Module. 

This module identifies mutually exclusive factual conflicts across the beliefs of parallel agents, employs a tool-based reasoning loop to resolve these conflicts, and applies concise, targeted updates to the agents’ beliefs to improve factual correctness. In contrast to post-hoc verification methods that operate on preliminary solutions, our module is integrated directly into the agent loop immediately after each execution step. By acting on the beliefs that govern future planning and execution, the module stabilizes agent reasoning before downstream decisions are made.

\begin{wraptable}{r}{0.46\textwidth}
\vspace{-0.15in}
\caption{
\textbf{Error Type Analysis.}
Error types in generation trajectories from four parallel Gemini-2.5-Flash-Lite agents, averaged over three critic models. Details in \Cref{sec:app-errortype}.
}
\centering\small

\begin{tabular}{@{}lcc@{}}
\toprule
Error Type & AIME & GAIA \\
\midrule
Common & \phantom{0}5.6\% & 13.6\% \\
Neutral & 22.9\% & 19.4\% \\
\textbf{Conflicting (Detectable)} & \textbf{71.5\%} & \textbf{67.0\%} \\
\bottomrule
\end{tabular}

\label{tab:error_type}
\end{wraptable}

\textbf{Observation.}
In \cref{tab:error_type}, we categorize errors in parallel agentic reasoning trajectories into three types. Specifically, we define errors as conflicting, common, or neutral based on whether other agents hold opposing beliefs, matching incorrect beliefs, or no relevant beliefs at the same reasoning step. Notably, 67.0\% to 71.5\% of errors are categorized as conflicting, suggesting that most errors are detectable by inspecting factual inconsistencies across multiple agents' beliefs. This observation naturally motivates our design of the consistency module.

\textbf{Conflict extraction.}
After parallel problem-solving agents $\{\mathcal{A}_i\}_{i=1}^{N}$ complete an execution step, the consistency module collects their beliefs $\{B_i\}_{i=1}^{N}$ and extracts conflicts, defined as sets of mutually exclusive factual claims. Each conflict $c$ is represented as a structured object containing the information described alongside. The module may also flag factual errors from individual agents, similar in spirit to self-reflection. A conflict instance is created for each set of mutually exclusive factual claims, and all conflicts are gathered into a conflict report $\mathcal{C}$.

\begin{wrapfigure}{r}{0.46\textwidth}
\vspace{-0.06in}
\begin{tcolorbox}[title=Structure of a Conflict, boxrule=0.5pt, left=4pt, right=4pt, top=2pt, bottom=2pt]
\label{box:conflict}
\begin{lstlisting}[language=yaml,basicstyle=\ttfamily\scriptsize]
conflict:
  agents: [0, 1, 2]
  description: |
    Brief description of the conflict
  claims:
    agent_0: Claim from agent 0
    agent_1: Claim from agent 1
    agent_2: Claim from agent 2
\end{lstlisting}
\end{tcolorbox}
\vspace{0.01in}

\begin{tcolorbox}[title=Structure of a Resolution, boxrule=0.5pt, left=4pt, right=4pt, top=2pt, bottom=2pt]
\label{box:resolution}
\begin{lstlisting}[language=yaml,basicstyle=\ttfamily\scriptsize]
resolution:
  agents: [0, 1, 2]
  description: |
    Brief description of the conflict
  claims:
    agent_0: Claim from agent 0
    agent_1: Claim from agent 1
    agent_2: Claim from agent 2
  correct_claim: |
    Claim verified by module
  justification: |
    How the claim was verified
\end{lstlisting}
\end{tcolorbox}
\vspace{-0.20in}
\end{wrapfigure}

\textbf{Conflict resolution via agentic reasoning.}
Given the conflict report $\mathcal{C}$, the consistency module initiates a dedicated ReAct loop to resolve the identified conflicts. The sole purpose of this loop is to determine the correct factual information for each conflict. Importantly, this conflict-resolution loop is decoupled from the parallel problem-solving agents, ensuring that no agent can directly access another agent's belief state, thereby preserving diversity. After resolution, the module produces a set of resolutions $\mathcal{R}$. 
Each resolution $r \in \mathcal{R}$ is transmitted only to the problem-solving agents originally involved in the conflict, as specified in the \texttt{agents} field. This design provides corrected information only to agents that require it, while preventing unrelated agents from being influenced by irrelevant information.

\textbf{Soft belief update.}
Once an agent $\mathcal{A}_i$ receives the resolutions associated with its index, it updates its belief $B_i$ accordingly. A straightforward approach is to directly replace conflicting belief entries with the resolved values, which we refer to as a hard update. However, conflict resolution is not guaranteed to be perfect, and incorrect resolutions may occasionally be produced. Blindly overwriting belief entries therefore risks introducing new factual errors that may propagate through subsequent reasoning steps. Such errors are particularly difficult to detect later, since multiple agents may share the same incorrect belief, reducing the likelihood of future conflicts. Moreover, even correct resolutions can reduce trajectory diversity if they repeatedly overwrite agent-specific beliefs with shared information.

To mitigate these risks, we adopt a soft update strategy. Each resolution is appended to $B_i$ rather than replacing the original conflicting information. The resolution is treated as an external suggestion that contradicts the existing belief and may itself be correct or incorrect. This preserves the agent's ability to recover from erroneous resolutions while encouraging it to re-examine the conflicting belief in subsequent reasoning. If the resolution is incorrect, the remaining inconsistency can be more easily detected by the agent later in the trajectory.

\subsection{Trajectory Diversification Module}
\label{sec:method-diversifier}

Maintaining diverse trajectories is critical for effective exploration in parallel agentic reasoning. It also supports the Consistency Module, since factual conflicts are easier to detect when agents pursue sufficiently distinct reasoning paths. However, agents initialized with identical prompts may suffer from trajectory collapse, especially when sharing information. To mitigate this, we introduce a Trajectory Diversification Module, which updates agent plans after each execution step to keep parallel trajectories sufficiently diverse.

\begin{wrapfigure}{r}{0.46\textwidth}
\vspace{-0.2in}
\begin{tcolorbox}[title=Structure of a Diversification Directive, boxrule=0.5pt, left=4pt, right=4pt, top=2pt, bottom=2pt]
\label{box:directive}
\begin{lstlisting}[language=yaml,basicstyle=\ttfamily\scriptsize]
directive:
  target_agent_index: |
    Index of the agent to modify plan
  modification_instruction: |
    Instruction specifying how to orthogonally shift the plan
\end{lstlisting}
\end{tcolorbox}
\vspace{-0.2in}
\end{wrapfigure}

\textbf{Batched analysis of plan-level redundancy.}
Similar to the Consistency Module, the diversification module adopts a centralized design. Given the set of plans $\{P_i\}_{i=1}^{N}$ from parallel agents $\{\mathcal{A}_i\}_{i=1}^{N}$, the module jointly analyzes all plans to identify redundancy and underexplored reasoning directions. Operating over this global plan configuration, the module produces a structured set of diversification directives $\mathcal{D}$. Specifically, it checks whether multiple trajectories are overly similar or whether plausible orthogonal reasoning paths remain unexplored. Each directive specifies how a particular agent's plan should be shifted to encourage exploration, and at most one directive is assigned to each agent at a given step.

This batched formulation ensures globally consistent diversification decisions and avoids oscillatory or conflicting local interventions. Once generated, directives are transmitted only to their corresponding agents. As with the consistency module, this targeted design prevents agents from inspecting other agents' plans, since they receive only the resulting directives rather than the full plans.

\textbf{Plan shifting via explicit instructions.}
Diversification directives are applied independently to the targeted agents. Unlike the soft belief update used in the consistency module, plan shifting is enforced once a directive is issued. Because different directives are assigned to different agents, this intervention does not risk inducing mode collapse. Rather, it is designed to explicitly promote diversity across trajectories. By confining intervention to the plan level while leaving task execution fully autonomous, the diversification module expands coverage of the solution space while preserving the coherence and interpretability of each reasoning trajectory.
\section{Experiments}
\label{sec:experiments}

In this section, we evaluate \mname across diverse reasoning benchmarks and analysis settings. We first describe the common experimental setup and baselines. In \Cref{sec:exp-benchmark}, we present main benchmark results comparing our method against standard test-time scaling strategies across multiple backbone models. We then analyze the sources of performance gains through error recovery analysis in \Cref{sec:experiment-recovery}. Next, we demonstrate the scalability of our framework in \Cref{sec:experiment-scaling} by scaling model size and the number of agents. In \Cref{sec:experiment-communication}, we compare \mname with other communication protocols. \Cref{sec:experiment-ablation} provides an ablation study examining the contributions of individual components and their effects on error recovery. Finally, in \Cref{sec:experiment-cost}, we discuss the efficiency of our approach.

\textbf{Common setup and baselines.}
All experiments are conducted within the Co-Sight framework, which we treat as an instantiation of the base agent loop described in \Cref{alg:base_agent}. We compare against the following baselines: Base Agent ($N = 1$), which performs a single pass without test-time scaling; Sequential Revision ($N = 1$), which introduces periodic self-revision; Independent Scaling, where multiple solver agents run in parallel without communication; Independent Scaling + Sequential Revision, which is a parallel version of Sequential Revision; Tree Search, implemented as beam search over reasoning trajectories; and \mname (ours), which augments the base loop with the Online Belief Consistency Module and the Trajectory Diversification Module.

Unless otherwise stated, we use $N = 4$ solver agents for all parallel scaling methods. We evaluate all methods on AIME 2024, AIME 2025, and GAIA, reporting majority-vote accuracy.

\begin{table*}[t]
\caption{\textbf{Benchmark Evaluations.} 
Comparison of test-time scaling methods on AIME 2024, AIME 2025, and GAIA. For AIME, agents have access to a Python code interpreter (C). For GAIA, agents have access to a Python code interpreter (C), file management tools (F), and web search tools (S). Parallel methods use four problem-solving agents. Due to cost constraints, GAIA L1/L2 is evaluated on a 50-sample subset. Scores are averaged over three runs, with the best values highlighted in \textbf{bold}.
}
\centering\small
\adjustbox{width=\linewidth}{

\begin{tabular}{@{}lccccccc@{}}
\toprule

 & N $>$ 1? & AIME 2024 & AIME 2025 & GAIA L1 & GAIA L2 & GAIA L3 & Average \\ 

\midrule
Available Tools & \multicolumn{1}{l}{} & C & C & C / F / S & C / F / S & C / F / S & \multicolumn{1}{l}{} \\

\midrule
\multicolumn{8}{c}{\cellcolor[HTML]{EFEFEF}\textit{Gemini-2.5-Flash-Lite}} \\
\midrule

Base Agent & \xmark & 61.4 & 46.1 & 35.8 & 21.5 & \phantom{0}5.9 & 34.1 \\
Sequential Revision (SR) & \xmark & 60.3 & 46.9 & 35.2 & 21.8 & \phantom{0}4.9 & 33.8 \\
Independent Scaling & \cmark & 65.3 & 52.2 & 44.0 & 28.5 & \phantom{0}9.0 & 39.8 \\
Independent Scaling + SR & \cmark & 65.3 & 54.2 & 42.7 & 29.2 & \phantom{0}5.6 & 39.4 \\
Tree Search & \cmark & 67.2 & 53.9 & 41.0 & 30.0 & 10.2 & 40.5 \\
\textbf{\mname (Ours)} & \cmark & \textbf{72.5} & \textbf{60.6} & \textbf{52.2} & \textbf{30.8} & \textbf{14.8} & \textbf{46.2} \\

\midrule
\multicolumn{8}{c}{\cellcolor[HTML]{EFEFEF}\textit{Qwen3.5-4B}} \\
\midrule

Base Agent & \xmark & 76.9 & 60.0 & 53.7 & 38.8 & 16.4 & 49.2 \\
Sequential Revision (SR) & \xmark & 78.6 & 60.6 & 54.0 & 36.7 & 21.0 & 50.2 \\
Independent Scaling & \cmark & 84.2 & 70.6 & 61.7 & 46.8 & 20.1 & 56.7 \\
Independent Scaling + SR & \cmark & 83.3 & 72.2 & 60.8 & 43.7 & 26.5 & 57.3 \\
Tree Search & \cmark & 78.3 & 62.8 & 58.7 & 45.3 & 16.7 & 52.4 \\
\textbf{\mname (Ours)} & \cmark & \textbf{87.8} & \textbf{75.8} & \textbf{69.5} & \textbf{49.8} & \textbf{28.7} & \textbf{62.3} \\ \bottomrule
\end{tabular}

}
\label{tab:main}
\end{table*}

\subsection{Main Results}
\label{sec:exp-benchmark}

This experiment evaluates whether \mname improves performance over commonly used test-time scaling strategies. We compare single solver, sequential self-revision, independent parallel scaling with and without self-revision, and tree search baselines against our method across diverse benchmarks spanning mathematical reasoning (AIME 2024, AIME 2025 \citep{balunovic2025aime}) and multi-level agentic reasoning (GAIA Levels 1-3 \citep{mialon2023gaia}). For reliability, all scores are averaged over three repeated runs.

As shown in \Cref{tab:main}, our method consistently outperforms the baselines across models and benchmarks, achieving average-score gains of 5.7\% and 5.0\% over the best-performing baseline. Independent scaling and tree search provide moderate improvements over a single solver. Sequential revision sometimes yields additional gains but remains limited by its reliance on error detection within a single reasoning trajectory. In contrast, our communicative approach achieves the strongest performance by combining the Consistency Module with trajectory diversification, mitigating error propagation and improving search-space diversity.

\subsection{Error Recovery Analysis}
\label{sec:experiment-recovery}

\begin{wraptable}{r}{0.5\textwidth}
\vspace{-0.15in}
\caption{
\textbf{Error Recovery Rates.}
Comparison of error recovery rates (\%) across benchmarks, defined as the percentage of factual or logical errors that are corrected during execution. Scores are averaged over three critic models. Detailed experiment setup is outlined in \Cref{sec:appendix-critic-model}.
}
\centering\small
\adjustbox{width=\linewidth}{

\begin{tabular}{@{}lcccccc@{}}
\toprule
 & \multicolumn{2}{c}{AIME} & \multicolumn{3}{c}{GAIA} & \multicolumn{1}{l}{} \\ 
\cmidrule(lr){2-3} \cmidrule(lr){4-6}
Method & 2024 & 2025 & L1 & L2 & L3 & Avg. \\

\midrule
\multicolumn{7}{c}{\cellcolor[HTML]{EFEFEF}\textit{Gemini-2.5-Flash-Lite} } \\
\midrule

Independent & 32.6 & 20.5 & 16.6 & 12.9 & \phantom{0}9.7 & 18.5 \\
Independent + SR & 28.7 & 25.2 & 18.6 & 12.8 & 14.4 & 19.9 \\
Tree Search & 22.5 & 10.8 & 12.3 & 10.6 & \phantom{0}8.4 & 12.9 \\
\textbf{\mname (Ours)} & \textbf{50.6} & \textbf{43.9} & \textbf{42.7} & \textbf{30.7} & \textbf{25.6} & \textbf{38.7} \\

\midrule
\multicolumn{7}{c}{\cellcolor[HTML]{EFEFEF}\textit{Qwen3.5-4B}} \\
\midrule

Independent & 67.1 & 59.6 & 43.3 & 38.5 & 30.7 & 47.8 \\
Independent + SR & 69.8 & 71.2 & 55.5 & 46.2 & 45.7 & 57.7 \\
Tree Search & 66.6 & 59.3 & 44.6 & 48.9 & 37.5 & 51.4 \\
\textbf{\mname (Ours)} & \textbf{70.7} & \textbf{76.3} & \textbf{61.7} & \textbf{53.7} & \textbf{46.5} & \textbf{61.8} \\ \bottomrule
\end{tabular}
}

\vspace{-0.15in}

\label{tab:error_recovery}
\end{wraptable}

To better understand why \mname outperforms independent and sequential baselines, we investigate the mechanisms of error detection and correction during execution by comparing the error recovery rates across different methods.

\textbf{Experimental Setup.}
Adopting the step-wise evaluation methodology from ProcessBench \citep{zheng2025processbenchidentifyingprocesserrors}, we compare agent execution results against ground-truth reference solutions to identify factual or logical errors and determine whether they are later resolved. Our primary metric is the Error Recovery Rate, which is the percentage of instances where an initial factual or logical error is successfully identified and corrected before the final answer is generated. To ensure a robust analysis, we employ a critic model to provide binary labels of step-wise correctness, following the critic model design from ProcessBench. Furthermore, we repeat the evaluation with three different critic models, namely Gemini-3-Flash \citep{pichai2025gemini3}, GPT-5-nano \citep{singh2025gpt5}, and Gemini-2.5-Flash-Lite \citep{comanici2025gemini25}, and take the average of their results. 

\textbf{Results.}
As shown in \Cref{tab:error_recovery}, \mname shows superior error detection and correction compared to baselines. \mname significantly outperforms all baselines, achieving average recovery rates of 38.7\% and 61.8\% for the Gemini and Qwen models, respectively, being consistently higher than the best-performing baselines.
Our framework achieves this resilience through its sophisticated communicative design, that allows agents to collaboratively identify and correct invalid reasoning steps that would otherwise propagate unchecked in isolated settings.

\subsection{Performance with Further Scaling}
\label{sec:experiment-scaling}

\begin{table}[t]

\caption{\textbf{Scaling experiments.}
We evaluate \mname under two scaling regimes: (a) model scaling with Qwen3-235B-A22B-Instruct and (b) agent scaling with $N=8$ parallel Gemini-2.5-Flash-Lite agents. We report majority-voting accuracy across AIME and GAIA, averaged over three runs. The best values are highlighted in \textbf{bold}, and runner-up values are {\ul underlined}.
}
\label{tab:scaling}

\begin{subtable}[h]{0.495\textwidth}
    \centering\small
    \caption{Model Scaling: Qwen3-235B-A22B-Instruct }
\adjustbox{width=\linewidth}{
\begin{tabular}{@{}lcccccc@{}}
\toprule
 & \multicolumn{2}{c}{AIME} & \multicolumn{3}{c}{GAIA} & \multicolumn{1}{l}{} \\ 
\cmidrule(lr){2-3} \cmidrule(lr){4-6}
Method & 2024 & 2025 & L1 & L2 & L3 & Avg. \\
\midrule
Base Agent & 80.6 & 69.7 & 53.7 & 39.7 & 21.0 & 52.9 \\
Seq. Revision & 80.8 & 69.2 & 53.2 & 39.3 & 23.8 & 53.3 \\
Independent & {\ul 89.2} & {\ul 76.4} & {\ul 60.8} & {\ul 46.5} & 22.8 & 59.1 \\
Independent + SR & 88.9 & 74.2 & 59.3 & 45.0 & \textbf{29.9} & {\ul 59.5} \\
Tree Search & 86.1 & 75.0 & 57.7 & 45.7 & 25.3 & 58.0 \\
\textbf{\mname (Ours)} & \textbf{89.7} & \textbf{83.1} & \textbf{62.7} & \textbf{50.2} & {\ul 29.3} & \textbf{63.0} \\ 
\bottomrule

\end{tabular}
}
\end{subtable}
\hfill
\begin{subtable}[h]{0.495\textwidth}
    \centering\small
    \caption{Agent Scaling: Gemini-2.5-Flash-Lite ($N=8$)}
\adjustbox{width=\linewidth}{
\begin{tabular}{@{}lcccccc@{}}
\toprule
 & \multicolumn{2}{c}{AIME} & \multicolumn{3}{c}{GAIA} & \multicolumn{1}{l}{} \\ 
\cmidrule(lr){2-3} \cmidrule(lr){4-6}
Method & 2024 & 2025 & L1 & L2 & L3 & Avg. \\
\midrule

Independent & {\ul 71.5} & 60.4 & 46.7 & {\ul 32.7} & {\ul 12.5} & {\ul 44.7} \\ 
Independent + SR & 69.4 & {\ul 61.5} & 48.1 & 32.2 & \phantom{0}9.6 & 44.2 \\
Tree Search & 67.8 & 53.9 & {\ul 48.6} & 30.9 & \phantom{0}9.3 & 42.1 \\
\textbf{\mname (Ours)} & \textbf{74.6} & \textbf{64.0} & \textbf{59.3} & \textbf{40.8} & \textbf{14.0} & \textbf{50.5} \\ \bottomrule
\\
\\
\end{tabular}
}
\end{subtable}
\end{table}

To evaluate the scalability of \mname, we extend the evaluation in two dimensions. First, we scale the parameter size of the base LLM to test whether the gains of our framework extend to larger models. To this end, we evaluate our approach on Qwen3-235B-A22B-Instruct \citep{qwen3techreport}. Second, we scale the number of solver agents to test whether the benefits of the communication protocol extend to more agents. Specifically, we double the agent count and evaluate Gemini-2.5-Flash-Lite with 8 parallel solver agents instead of 4. As shown in \Cref{tab:scaling}, \mname consistently outperforms the baselines in both scaling regimes, demonstrating the scalability of our approach. This suggests that the effectiveness of our approach is complementary to underlying model capacity rather than specific to a particular model size or number of agents. 

\subsection{Comparison Against Other Communication Protocols}
\label{sec:experiment-communication}

\begin{wraptable}{r}{0.5\textwidth}
\vspace{-0.15in}
\caption{
\textbf{Communication Baselines.}
Comparison of \mname against Multi-Agent Debate and Mixture-of-Agents, adapted to the exploration-stage agent loop using Gemini-2.5-Flash-Lite. We report majority-voting accuracy and diversity scores, defined as $100-\text{Self-BLEU}$.
}
\centering\small

\adjustbox{width=\linewidth}{

\begin{tabular}{@{}lcccccc@{}}
\toprule
 & \multicolumn{2}{c}{AIME} & \multicolumn{3}{c}{GAIA} & \\ \cmidrule(lr){2-3} \cmidrule(lr){4-6}
Method & 2024 & 2025 & L1 & L2 & L3 & Avg. \\

\midrule
\multicolumn{7}{c}{\cellcolor[HTML]{EFEFEF}\textit{Majority Voting Accuracy}} \\
\midrule

Multi-Agent Debate & 64.4 & 52.8 & 49.3 & 30.0 & \phantom{0}8.3 & 41.0 \\
Mixture-of-Agents & 63.9 & 52.5 & 51.3 & \textbf{34.0} & 11.1 & 42.6 \\
\textbf{\mname (Ours)} & \textbf{72.5} & \textbf{60.6} & \textbf{52.2} & 30.8 & \textbf{14.8} & \textbf{46.2} \\

\midrule
\multicolumn{7}{c}{\cellcolor[HTML]{EFEFEF}\textit{ Diversity Score (Trajectory-Level)}} \\
\midrule

Multi-Agent Debate & 53.3 & 54.7 & 51.1 & 51.5 & 54.2 & 53.0 \\
Mixture-of-Agents & \phantom{0}2.8 & \phantom{0}2.5 & \phantom{0}5.4 & \phantom{0}5.7 & \phantom{0}4.5 & \phantom{0}4.2 \\
\textbf{\mname (Ours)} & \textbf{63.1} & \textbf{62.1} & \textbf{59.7} & \textbf{58.7} & \textbf{59.1} & \textbf{60.5} \\

\midrule
\multicolumn{7}{c}{\cellcolor[HTML]{EFEFEF}\textit{ Diversity Score (Step-Level)}} \\
\midrule

Multi-Agent Debate & 46.2 & 45.4 & 42.3 & 42.3 & 44.8 & 44.2 \\
Mixture-of-Agents & 10.9 & 11.3 & 13.3 & 13.3 & 11.8 & 12.1 \\
\textbf{\mname (Ours)} & \textbf{55.4} & \textbf{52.6} & \textbf{51.0} & \textbf{48.2} & \textbf{49.1} & \textbf{51.3} \\ 

\bottomrule

\end{tabular}

}

\label{tab:mad}
\vspace{-0.1in}
\end{wraptable}

Many representative communication protocols, including Multi-Agent Debate \citep{du2024mad} and Mixture-of-Agents \citep{wang2024moa}, are primarily designed for \textit{solution-stage refinement}. In this setup, agents exchange, critique, or aggregate preliminary answers and the accompanying reasoning trajectory to improve candidate solutions. In contrast, \mname targets a different problem: \textit{exploration-stage error control}. Rather than intervening after preliminary answers have already formed, \mname intervenes during the agent execution loop to detect factual conflicts, resolve them before they propagate, and preserve trajectory diversity for subsequent reasoning. 

In this section, we compare \mname with two representative communication protocols, namely Multi-Agent Debate and Mixture-of-Agents. In Multi-Agent Debate, agents repeatedly share their progress with one another and revise their reasoning based on peer responses. In Mixture-of-Agents, agents' outputs are centrally aggregated at each synchronization step, and subsequent reasoning proceeds from the aggregated state. These methods were originally designed for iterative refinement of candidate solutions, rather than exploration-stage communication within an agent execution loop. We therefore adapt them to this setting as strong communication baselines, allowing us to test whether generic information exchange is sufficient or whether the targeted design of \mname is necessary for reducing error propagation.

As shown in \Cref{tab:mad}, \mname consistently outperforms both protocols across benchmarks. This gap reflects the different objective of our communication design. Multi-Agent Debate and Mixture-of-Agents exchange broad intermediate information, which can improve consensus but may also cause premature synchronization of reasoning trajectories. In contrast, \mname uses targeted communication, sharing only conflict resolutions with the agents involved and applies explicit trajectory diversification to maintain independent exploration. This allows \mname to reduce error propagation without collapsing agents into a shared reasoning path.

This distinction is further supported by the diversity analysis. Following prior work \citep{liang2024encouragingdivergentthinkinglarge,zhang2024improvingdiversitycommonsensegeneration}, we adopt $100 - \text{Self-BLEU}$ to measure semantic diversity among agent trajectories. In addition to trajectory-level diversity, we measure step-level diversity by comparing the reasoning traces of parallel agents at the same step. For both metrics, \mname consistently shows higher trajectory diversity across benchmarks, indicating that exploration-stage communication can improve reliability while preserving diverse reasoning paths.

\subsection{Ablation and Analysis}
\label{sec:experiment-ablation}

\begin{wraptable}{r}{0.5\textwidth}
\vspace{-0.15in}
\caption{
\textbf{Component Ablation.}
Ablation of \mname on AIME 2024 and GAIA Level 3 using Gemini-2.5-Flash-Lite. We report majority-voting accuracy and error recovery rates. Soft belief updates improve belief consistency by avoiding rigid synchronization, while trajectory diversification yields the largest gains, especially on GAIA.
}
\centering\small
\vspace{0.1in}
\adjustbox{width=\linewidth}{

\begin{tabular}{@{}lcc@{}}
\toprule
 & AIME 2024 & GAIA Level 3 \\ 
 
\midrule
\multicolumn{3}{c}{\cellcolor[HTML]{EFEFEF}\textit{Majority Voting Accuracy}} \\
\midrule

Independent & 65.3 & \phantom{0}9.0 \\
\quad + Consistency Module & 68.3 & \phantom{0}8.0 \\
\quad + Soft Belief Updates & 71.4 & \phantom{0}9.3 \\
\quad + Diversification Module & \textbf{72.5} & \textbf{14.8} \\

\midrule
\multicolumn{3}{c}{\cellcolor[HTML]{EFEFEF}\textit{Error Recovery Rates}} \\
\midrule

Independent & 32.6 & \phantom{0}9.7 \\
\quad + Consistency Module & 33.5 & 12.0 \\
\quad + Soft Belief Updates & 33.6 & 13.3 \\
\quad + Diversification Module & \textbf{50.6} & \textbf{25.6} \\ \bottomrule
\end{tabular}

}

\label{tab:ablation}
\vspace{-0.2in}
\end{wraptable}

In this section, we analyze \mname to isolate the contributions of the Consistency Module, soft belief updates, and trajectory diversification. Unless otherwise specified, we use $N=4$ solver agents across all methods in this section.

As shown in \Cref{tab:ablation}, adding the Consistency Module without soft belief updates slightly degrades GAIA Level 3 accuracy, suggesting that conflict resolution alone may introduce harmful synchronization in more open-ended search spaces. Soft belief updates mitigate this issue by appending resolutions as external suggestions rather than overwriting agent beliefs. This improves performance on both benchmarks, indicating that agents benefit from peer-derived corrections while retaining their independent reasoning trajectories.

Finally, adding the Diversification Module yields the strongest performance, with particularly large gains on GAIA Level 3 and error recovery rates. This suggests that diversification is essential in large search spaces, where agents must explore complementary reasoning paths rather than prematurely converge on a single trajectory. Overall, both components are necessary. Belief consistency improves error detection and correction, while diversification preserves broad search-space coverage and substantially improves recovery from earlier errors.

\subsection{Cost Analysis}
\label{sec:experiment-cost}

\begin{wrapfigure}{r}{0.46\textwidth}
\centering\small
\vspace{-0.2in}
\includegraphics[width=0.95\linewidth]{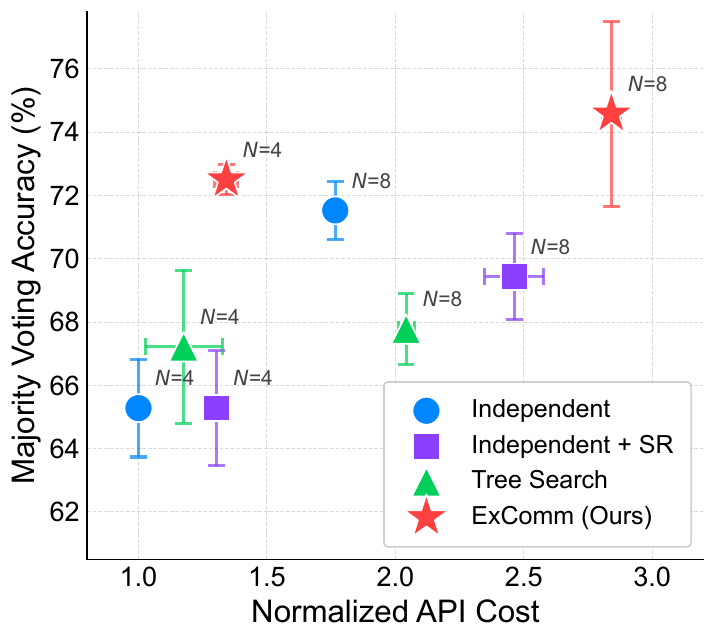}
\caption{
\textbf{Performance-Cost Trade-Off.} 
We plot majority-voting accuracy versus normalized API cost on AIME 2024 using Gemini-2.5-Flash-Lite.
Error bars are standard errors.
}
\label{fig:efficiency}
\vspace{-0.3in}
\end{wrapfigure}

To assess the efficiency of \mname, we analyze the per-sample API cost of each method and summarize the performance-cost trade-off in \Cref{fig:efficiency}. \mname achieves the best trade-off, matching or outperforming the $N=8$ baselines with only $N=4$ solver agents while requiring lower API cost. This suggests that targeted exploration-stage communication can be more cost-effective than simply increasing the number of independent trajectories. We also report latency in \Cref{tab:efficiency}. For example, \mname has a similar per-sample latency to the $N=8$ Independent + SR baseline (10.45 min. vs. 9.57 min.), 
while requiring significantly less API cost
and improving accuracy on both AIME and GAIA. Overall, \mname achieves a favorable performance-cost trade-off while maintaining latency in a comparable range to parallel baselines.

\section{Conclusion}
\label{sec:conclusion}
We introduce \mname, a communication protocol for the exploration stage of agentic test-time scaling. By auditing agents' beliefs for factual inconsistencies and coordinating plan-level diversification, \mname directly addresses a key failure mode of agentic systems, specifically the propagation of incorrect intermediate beliefs that corrupt downstream reasoning. Our results suggest that the design space of exploration-stage communication (what to share, when to intervene, and how to balance correction with diversity) is a promising and largely unexplored axis for scaling agentic reasoning. As demonstrated in the results, targeted interventions in this space yield substantial gains over both independent scaling and adapted solution-stage communication protocols.

\bibliography{main}
\bibliographystyle{unsrtnat}

\newpage
\appendix
\onecolumn

\section{Further Discussions}
\label{sec:appendix-impact}

\subsection{Broader Impact Statement}

This work studies methods for improving the reliability of agentic test-time scaling through exploration-stage communication and error control. By detecting belief-level inconsistencies among parallel solver agents and resolving them before they propagate, \mname may contribute to more robust language-model deployment in complex reasoning settings. The proposed framework also aims to improve the performance-cost trade-off by using targeted coordination rather than simply increasing the number of independent trajectories.

At the same time, 
more capable automated reasoning systems may also be misused or over-trusted, especially in high-stakes domains where incorrect outputs can have significant consequences. As with other advances in language-model reasoning, these techniques should be applied responsibly, with appropriate human oversight, domain-specific validation, and consideration of their broader societal impacts.

\subsection{Limitations and Future Work}

Our work focuses on error propagation during the exploration stage of agentic test-time scaling, where agents are still forming reasoning trajectories through iterative reasoning and tool use. This scope is intentional, as early belief-level errors can influence many downstream decisions. However, our method is not designed to replace aggregation or refinement after candidate answers are generated.

Another consideration is the quality of the consistency module. \mname uses a dedicated agentic verifier to resolve factual conflicts, and its effectiveness depends on the verifier's ability to produce reliable corrections. Since incorrect resolutions may introduce new errors, \mname uses soft belief updates that present corrections as external suggestions rather than replacing the original beliefs. Empirically, \mname improves performance across all tested models, but in extreme cases, repeated incorrect verification results could still mislead agents and degrade performance. Improving verifier reliability and uncertainty estimation is therefore a promising direction for making exploration-stage communication more robust.

\section{Implementation Details on Critic Model}
\label{sec:appendix-critic-model}

\subsection{Design of Critic Model}

The Critic Model framework, as introduced in ProcessBench \cite{zheng2025processbenchidentifyingprocesserrors}, was originally employed in the mathematical domain to evaluate candidate solutions for a given problem and identify the first occurence of an error. They reported that critic models utilizing proprietary language models (e.g., o1-mini) with simple prompting achieved high agreement with human annotations. Notably, these models were able to not only correctly localize the erroneous step but also provide a detailed thinking process and explanation.

Building upon this foundation, we adapt the ProcessBench prompt template to suit our specific objectives. We introduce two primary modifications:

\begin{enumerate}
    \item \textbf{Additional Use of Reference Solutions.} Unlike ProcessBench, which aimed to benchmark the performance of the critic itself, our objective is to maximize the accuracy of the critique. To this end, we provide the critic with a ground-truth reference solution. This is particularly beneficial for tool-augmented benchmarks like GAIA; without a reference solution containing necessary intermediate results (e.g., outputs from web searches), the critic is prone to misjudgment due to missing context.

    \item \textbf{Identification of Error Recovery.} While ProcessBench focused solely on detecting the existence and location of the first error, we extend this scope to determine if the error is also recovered in subsequent steps. Although detecting error recovery is a more complex task, the provision of a reference solution significantly mitigates the difficulty. By treating the reference solution---which contains key intermediate results---as a roadmap, the model can effectively compare execution traces and verify if the reasoning trajectory realigns with the correct path.
\end{enumerate}

\begin{figure}[t!]
    \centering
    \begin{tcolorbox}[colback=gray!5!white,colframe=gray!50!black,title=Critic Model Prompt Template, fonttitle=\bfseries\small]
    \footnotesize
    \begin{verbatim}
You are an expert Critic. Your goal is to evaluate the solving process
of a specific agent ({agent_id}) regarding the given problem.

# Problem: {problem}

# Reference Solution: {reference_solution}
# Reference Answer: {reference_answer}

# Model Execution of {agent_id}
The execution log below belongs to {agent_id}. Focus on {agent_id}'s workflow only.
<model_execution>
  <step_0>
    Action: {Execution results of step 0}
    Memory Diff: {Updated Intermediate Results of step 0}
  </step_0>
  ...
  <step_K>
    Action: {Execution results of step K}
    Memory Diff: {Updated Intermediate Results of step K}
  </step_K>
</model_execution>

# Your Task
1. Analyze the "Action" and "Memory Diff" of {agent_id}.
2. Identify "Factual Errors" committed by {agent_id}.
   - Definition: A Factual Error is where {agent_id} incorrectly derives
     an intermediate result or maintains an incorrect intermediate result
     that impacts decision-making.
   - Exclusions: Minor tool errors, or errors made by other agents.

3. For each Factual Error:
   - Error Type: Description.
   - Occurrence Step: Step number where {agent_id} introduced the error.
   - Recovered Step: Step number where the error was corrected (or "N/A").

4. Final Check:
   - If the agent's final answer is INCORRECT (differs from Reference Answer),
     there MUST be at least one Factual Error that is "N/A" (Unrecovered).
   - Do NOT mark an error as "Recovered" if the agent proceeded to a wrong
     conclusion based on a related misconception.

# Output Format
Provide your analysis in the following XML format:
<error_recovery_analysis>
  <error>
    <description>...</description>
    <occurrence_step>...</occurrence_step>
    <recovered_step>...</recovered_step>
  </error>
</error_recovery_analysis>
    \end{verbatim}
    \end{tcolorbox}
    \caption{The prompt template used for the critic model.}
    \label{fig:critic-prompt}
\end{figure}

\clearpage

\subsection{Prompt Template}
\label{sec:prompt-template}

Our critic model utilizes a structured prompt designed to detect factual errors and their subsequent recovery. The prompt (see \Cref{fig:critic-prompt}) inputs include the problem statement, a ground-truth reference solution, and the target agent's step-by-step execution log. The critic is instructed to identify ``Factual Errors'', defined as incorrect derivations impacting decision-making, while ignoring minor tool slips or external noise. Crucially, the prompt enforces a temporal analysis: for each error, the critic must pinpoint the \textit{Occurrence Step} and, if applicable, the \textit{Recovered Step}. To ensure consistency, the instructions mandate that any trajectory resulting in an incorrect final answer must be associated with at least one unrecovered error.

\section{Error Type Analysis}
\label{sec:app-errortype}

This section describes the experimental setup for the error type classification reported in \Cref{tab:error_type}. The goal is to determine, for each factual error in independent-baseline trajectories, whether the error would be detectable through cross-agent conflict analysis.

\textbf{Error identification.}
We use an LLM-as-a-Judge approach to identify factual errors in each agent's execution trajectory. The critic model setup follows the same design described in \Cref{sec:appendix-critic-model}. Given a problem, a reference solution, and an agent's step-by-step execution log, the critic identifies factual errors and the steps at which they occur. For each identified error, we then classify whether the error would be detectable by examining the beliefs of other parallel agents at the corresponding step.

\textbf{Classification procedure.}
For each identified error, we examine the reasoning logs of the other three parallel agents at the same execution step. A critic model classifies the error into one of the following categories based on the other agents' beliefs at the time of the error:

\begin{itemize}[leftmargin=1.5em, itemsep=2pt]
\item \textbf{Conflicting:} At least one other agent has established a different factual claim on the same topic, whether correct or incorrect. The presence of a conflicting claim makes the error detectable via cross-agent comparison.
\item \textbf{Common:} All other agents make the same or an equivalent error. No conflicting claim exists, so the error is not detectable through cross-agent conflict analysis.
\item \textbf{Neutral:} No other agent explicitly discusses or contradicts this specific fact in their completed reasoning at the given step. The error goes unmentioned.
\end{itemize}

To ensure temporal fairness, each other agent's reasoning log is truncated to include only steps up to and including the step at which the error occurs. This prevents the classification from relying on information that would not yet be available at the time the error is committed.

\begin{figure}[t!]
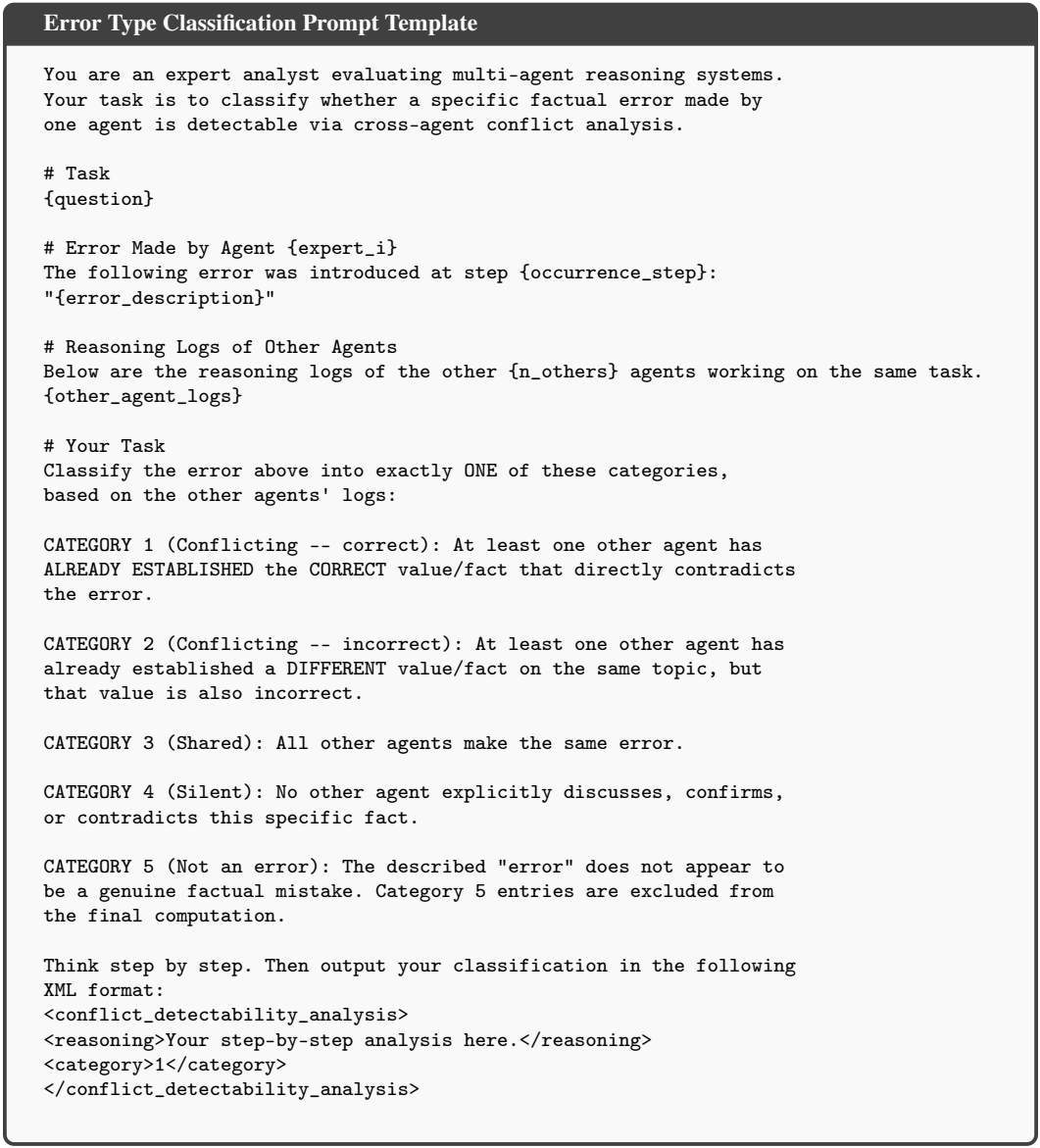

    \centering
    \begin{tcolorbox}[colback=gray!5!white,colframe=gray!50!black,title=Error Type Classification Prompt Template, fonttitle=\bfseries\small]
    \footnotesize
    \begin{verbatim}
You are an expert analyst evaluating multi-agent reasoning systems.
Your task is to classify whether a specific factual error made by
one agent is detectable via cross-agent conflict analysis.

# Task
{question}

# Error Made by Agent {expert_i}
The following error was introduced at step {occurrence_step}: 
"{error_description}"

# Reasoning Logs of Other Agents
Below are the reasoning logs of the other {n_others} agents working on the same task.
{other_agent_logs}

# Your Task
Classify the error above into exactly ONE of these categories,
based on the other agents' logs:

CATEGORY 1 (Conflicting -- correct): At least one other agent has
ALREADY ESTABLISHED the CORRECT value/fact that directly contradicts
the error.

CATEGORY 2 (Conflicting -- incorrect): At least one other agent has
already established a DIFFERENT value/fact on the same topic, but
that value is also incorrect.

CATEGORY 3 (Shared): All other agents make the same error.

CATEGORY 4 (Silent): No other agent explicitly discusses, confirms,
or contradicts this specific fact.

CATEGORY 5 (Not an error): The described "error" does not appear to
be a genuine factual mistake. Category 5 entries are excluded from
the final computation.

Think step by step. Then output your classification in the following
XML format:
<conflict_detectability_analysis>
<reasoning>Your step-by-step analysis here.</reasoning>
<category>1</category>
</conflict_detectability_analysis>
    \end{verbatim}
    \end{tcolorbox}
    \caption{Prompt template for error type classification. Categories~1 and~2 are merged as ``Conflicting'' in \Cref{tab:error_type}. Category~3 maps to ``Common''. Category~4 maps to ``Neutral''}
    \label{fig:detectability-prompt}
\end{figure}

\textbf{Evaluation.}
We run the classification using three critic models (Gemini-3-Flash, GPT-5-nano, and Gemini-2.5-Flash-Lite) and report the average across all three. The analysis is conducted on errors identified from four parallel Gemini-2.5-Flash-Lite agents running Independent scaling on AIME 2024, AIME 2025, and GAIA. The prompt template used for the classification is shown in \Cref{fig:detectability-prompt}.

\section{Latency Analysis}
\label{sec:appendix-latency}

\Cref{tab:efficiency} reports the per-sample latency of each method. The additional latency of \mname relative to Independent scaling mainly comes from the Consistency Module, which runs a separate tool-augmented reasoning loop, along with the plan analysis performed by the Trajectory Diversification Module. Despite this overhead, \mname at $N=4$ outperforms all $N=8$ baselines on both AIME and GAIA, suggesting that targeted coordination uses each trajectory more effectively.


\begin{table}[!ht]
\caption{
\textbf{Latency Analysis.} We report median per-sample latency (minutes) with Gemini-2.5-Flash-Lite. Accuracy columns show the average over AIME 2024/2025 and GAIA L1/L2/L3, respectively. \mname at $N=4$ outperforms scaled-up baselines at $N=8$ while maintaining modest additional latency.
}
\centering\small
\begin{tabular}{@{}lcccc@{}}
\toprule
 & N & Latency (min.) & Accuracy (AIME) & Accuracy (GAIA) \\ \midrule
Independent & 8 & \phantom{1}7.94 & 66.0 & 30.6 \\
Independent + SR & 8 & \phantom{1}9.57 & 65.5 & 30.0 \\
Tree Search & 8 & \phantom{1}7.38 & 60.8 & 29.6 \\
\textbf{\mname (Ours)} & 4 & 10.45 & \textbf{66.5} & \textbf{32.6} \\ \bottomrule
\end{tabular}
\label{tab:efficiency}
\end{table}

\clearpage

\section{Example Illustrations}
\label{sec:appendix-examples}

This section provides concrete examples of the structured objects used by the Consistency Module and the Trajectory Diversification Module, illustrating the format of conflicts, resolutions, and diversification directives introduced in \Cref{sec:method}.

\subsection{Conflict and Resolution Objects}

A \emph{conflict} is a structured object that identifies mutually exclusive factual claims held by different agents at the same reasoning step. The \texttt{claims} field records each involved agent's specific claim on the contested topic. A \emph{resolution} is the output of the tool-augmented verification loop that adjudicates the conflict. It contains the verified correct claim, a justification, and the list of agents that should receive the correction.

As a concrete illustration, consider a scenario where agents need to identify the largest two-digit number divisible by both 6 and 8 as an intermediate step. Agent~1 correctly computes 96 (noting $\text{LCM}(6,8)=24$ and $24 \times 4 = 96$). Agent~2 stops at 72 ($24 \times 3 = 72$, having not searched far enough). The Consistency Module flags this disagreement and extracts the following conflict:

\begin{center}
\begin{minipage}{0.65\textwidth}
\begin{tcolorbox}[title=Example Conflict, boxrule=0.5pt, left=4pt, right=4pt, top=2pt, bottom=2pt]
\begin{lstlisting}[language=yaml,basicstyle=\ttfamily\scriptsize]
conflict:
  agents: [1, 2]
  description: |
    Agents disagree on the largest two-digit number
    divisible by both 6 and 8. Agent 1 identifies 96;
    Agent 2 identifies 72.
  claims:
    agent_1: "96"
    agent_2: "72"
\end{lstlisting}
\end{tcolorbox}
\end{minipage}
\end{center}

The Consistency Module then invokes a tool-augmented ReAct loop that tests both candidates via the Python interpreter, without assuming either agent is correct. It verifies divisibility of both 96 and 72 by 6 and 8, confirms both are valid multiples, and selects the larger one:

\begin{center}
\begin{minipage}{0.65\textwidth}
\begin{tcolorbox}[title=Example Resolution, boxrule=0.5pt, left=4pt, right=4pt, top=2pt, bottom=2pt]
\begin{lstlisting}[language=yaml,basicstyle=\ttfamily\scriptsize]
resolution:
  agents: [1, 2]
  description: |
    Agents disagree on the largest two-digit number
    divisible by both 6 and 8.
  claims:
    agent_1: "96"
    agent_2: "72"
  correct_claim: "96"
  reason: |
    Both 96 and 72 are divisible by 6 and 8 (verified
    by Python). Since 96 > 72, 96 is the largest.
    Agent 2 stopped its search prematurely.
\end{lstlisting}
\end{tcolorbox}
\end{minipage}
\end{center}

This resolution is then incorporated into the involved agents' belief states via soft updates, appending the verified information as an external suggestion rather than overwriting existing beliefs.

\subsection{Diversification Directive}

Suppose Agents~0 and~1 both plan to solve the same sub-problem by iterating downward from 100 and checking whether each number is divisible by both 6 and 8. The Trajectory Diversification Module detects this redundancy and issues a directive to redirect one agent toward an orthogonal approach:

\begin{center}
\begin{minipage}{0.68\textwidth}
\begin{tcolorbox}[title=Example Diversification Directive, boxrule=0.5pt, left=4pt, right=4pt, top=2pt, bottom=2pt]
\begin{lstlisting}[language=yaml,basicstyle=\ttfamily\scriptsize]
directive:
  target_agent_index: 1
  modification_instruction: |
    Instead of iterating downward from 100, compute the
    set of all positive integers less than 100 that are
    divisible by 6, then filter for those also divisible
    by 8, and return the maximum of the resulting set.
\end{lstlisting}
\end{tcolorbox}
\end{minipage}
\end{center}

This shifts Agent~1 from a sequential brute-force search to a set-based filtering approach, ensuring that the two agents explore complementary strategies for the same sub-problem.

\section{{Case Studies}}
\label{sec:appendix-case-studies}

{We present representative case studies from evaluation trajectories to illustrate how the Consistency Module and the Trajectory Diversification Module operate in practice.}

{\textbf{Consistency Module (AIME 2024 \#19).}
An agent incorrectly modeled a geometric locus as a circle rather than an astroid. The Consistency Module flagged the geometric conflict across agents, verified the astroid equation via tool-augmented reasoning, and issued a soft update. The failing agent recovered and adopted the correct geometric model without losing its prior reasoning context.}

{\textbf{Diversification Module (AIME 2025-II \#6).}
Agent plans initially converged on coordinate geometry, creating a shared structural bottleneck. The Trajectory Diversification Module detected this redundancy and directed one agent to pivot to synthetic geometry. This produced an orthogonal trajectory that uniquely solved the task.}

{\textbf{Soft update robustness (GAIA e1fc63a2).}
The Consistency Module erroneously corrected a valid calculation due to unit confusion. Because corrections are applied as soft updates rather than hard overwrites, agents retained their valid derivations alongside the flawed suggestion. They recognized the verifier's error in subsequent reasoning steps and answered correctly. This case illustrates that the soft update design is necessary for robustness against imperfect conflict resolution.}

\end{document}